\documentclass[10pt,twocolumn,letterpaper]{article}

\usepackage{iccv}              
\usepackage{amsmath,amssymb,amsfonts}
\usepackage{algorithm,algorithmic} 
\usepackage{graphicx}
\usepackage{textcomp}
\usepackage{xcolor}
\usepackage{diagbox}
\usepackage{makecell}
\usepackage{subcaption}
\usepackage{diagbox}  
\usepackage{booktabs} 
\usepackage{multirow} 
\usepackage{adjustbox} 

\definecolor{iccvblue}{rgb}{0.21,0.49,0.74}
\usepackage[pagebackref,breaklinks,colorlinks,allcolors=iccvblue]{hyperref}

\def\paperID{15372} 
\def\confName{ICCV}
\def\confYear{2025}

\title{InSPE: Rapid Evaluation of Heterogeneous Multi-Modal \\ Infrastructure Sensor Placement }

\author{
Zhaoliang Zheng$^{\star}$\thanks{$^{\star}$Equal contribution. $^{\dagger}$Corresponding authors: jiaqima@ucla.edu; \\ meng925@g.ucla.edu},
Yun Zhang$^{\star}$, 
Zonglin Meng$^{\dagger}$, 
Johnson Liu,
Xin Xia, 
Jiaqi Ma$^{\dagger}$ \\
University of California, Los Angeles
}

\begin{document}

\maketitle
\begin{abstract}

Infrastructure sensing is vital for traffic monitoring at safety hotspots (e.g., intersections) and serves as the backbone of cooperative perception in autonomous driving. While vehicle sensing has been extensively studied, infrastructure sensing has received little attention, especially given the unique challenges of diverse intersection geometries, complex occlusions, varying traffic conditions, and ambient environments like lighting and weather.
To address these issues and ensure cost-effective sensor placement, we propose Heterogeneous Multi-Modal Infrastructure Sensor Placement Evaluation (InSPE), a perception surrogate metric set that rapidly assesses perception effectiveness across diverse infrastructure and environmental scenarios with combinations of multi-modal sensors. InSPE systematically evaluates perception capabilities by integrating three carefully designed metrics, i.e.,  sensor coverage, perception occlusion, and information gain. To support large-scale evaluation, we develop a data generation tool within the CARLA simulator and also introduce Infra-Set, a dataset covering diverse intersection types and environmental conditions. Benchmarking experiments with state-of-the-art perception algorithms demonstrate that InSPE enables efficient and scalable sensor placement analysis, providing a robust solution for optimizing intelligent intersection infrastructure.

\end{abstract}    
\section{Introduction}

\begin{figure}[t]
\centering
\includegraphics[width=0.49\textwidth]{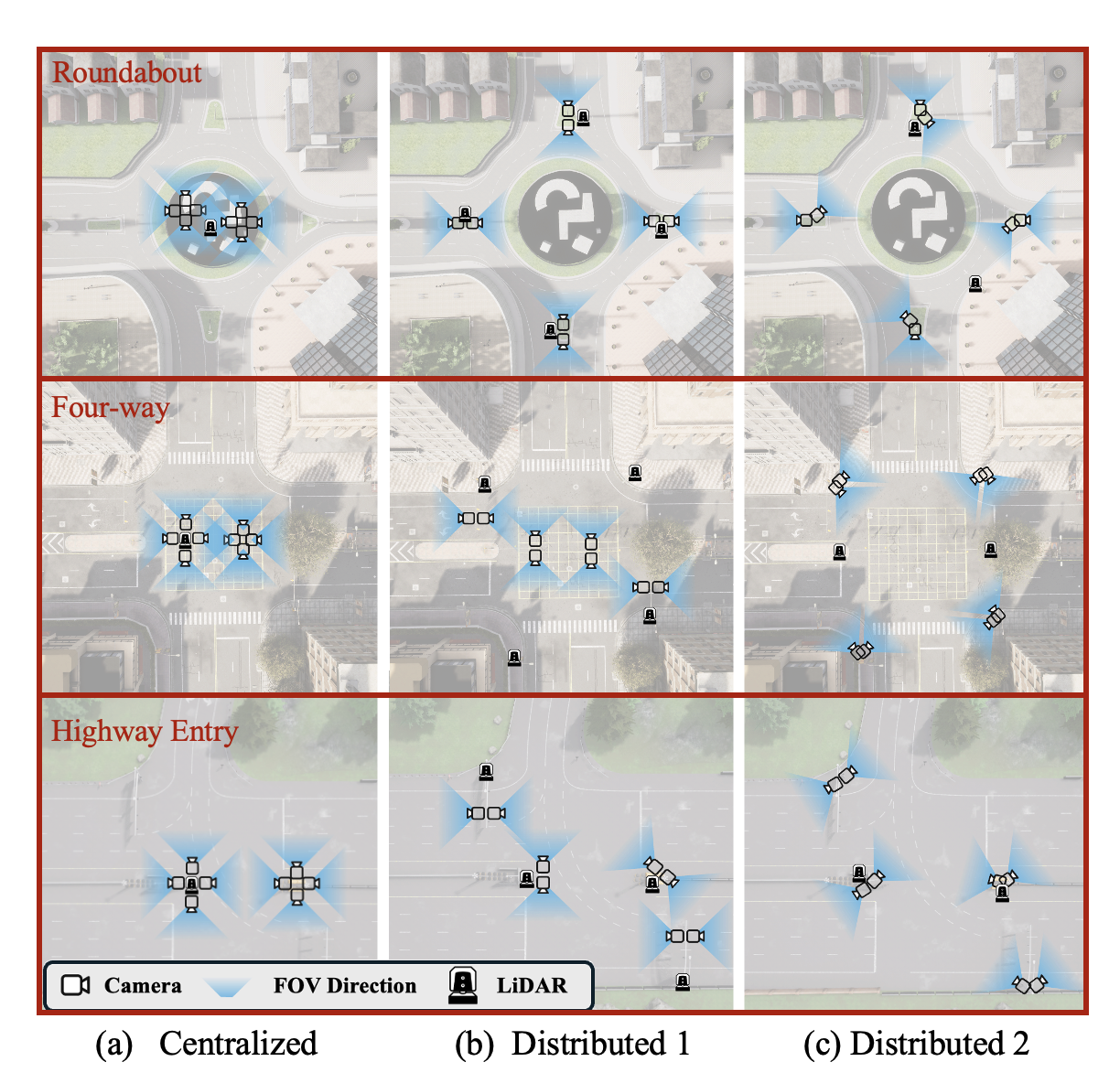}
\caption{Illustration figure of three types of sensor placements at three example intersections. (a) features sensors concentrated near the center of the intersection, whereas (b) and (c) employ a more dispersed placement throughout the intersection. The camera arrangement in (a) is similar to that of the V2XSet \cite{xu2022v2xvit} dataset, (b) resembles those in the DAIR-V2X \cite{yu2022dairv2x} and RCooper \cite{hao2024rcooper} datasets, and (c) is akin to the V2X-Real  \cite{xiang2024v2xreal} dataset. FOV direction is the field of view direction of the camera.}
\label{fig:sensor_placement}
\vspace{-6mm}
\end{figure}


Infrastructure sensing plays a crucial role in monitoring safety-critical intersections, not only enhancing traditional traffic management, but also serving as the foundation for connected vehicles and cooperative perception in autonomous driving \cite{ITE_RSU_Std_v1,ITE_RSU_SDR}. In this paper, we define a modern intelligent Infrastructure Unit (IU) that includes Roadside Units (RSUs) for wireless communication and multi-modal sensors for environmental perception, facilitating seamless data exchange and robust situational awareness. Unlike autonomous vehicles, which operate dynamically across varying locations and rely on lower-altitude sensors susceptible to significant occlusion, infrastructure-based sensors can be strategically positioned within a relatively stable roadway environment. Their strategic deployment—such as elevated placement or positioning that avoids occlusions from various road users—enhances their effectiveness \cite{xu2022v2xvit,yu2022dairv2x,hao2024rcooper,xiang2024v2xreal}. Specifically, strategic installation at intersections enables these infrastructure sensors to achieve broader coverage and significantly reduce occlusion, resulting in more reliable and comprehensive environmental perception compared to vehicle-mounted sensors \cite{guerna2022roadside,barrachina2013road,bhover2017v2x}. 

Additionally, infrastructure sensor deployment must carefully consider cost constraints, given the extensive number of intersections and safety-critical locations requiring coverage. High-end LiDAR sensors, such as those with 128 scan lines, offer precise and dense data within intersections but often yield sparse point clouds upstream of stop bars, thereby limiting analytical capabilities in these critical approach zones. Alternatively, deploying a combination of cameras and multiple lower-end LiDAR units (or equivalent sensor technologies) may provide a more balanced solution. Although the accuracy of combined sensor data may be slightly compromised in specific regions, such deployments can deliver broader spatial coverage and enhanced analytical flexibility at a more feasible cost. Therefore, it is essential to achieve an optimal balance between sensor performance and economic feasibility when designing infrastructure-based sensing systems.

Previous work on intersection sensor placement has primarily focused on LiDAR-only setups \cite{qu2023seip,kim2023placement,jiang2023optimizing}, overlooking the advantages of multi-modal sensing. These studies analyze LiDAR placement solely on point cloud distributions at intersection junctions \cite{cai2211analyzing} without leveraging latent information, such as vector maps, and neglecting key factors, such as occlusion and spatial coverage across diverse intersection geometries.  
Moreover, intersection geometries, road conditions, and infrastructure vary significantly, as illustrated in Fig.\ref{fig:sensor_placement}. Sensor placement strategies must be carefully designed to maximize the system's perception capabilities. Existing approaches include centralized camera placement, as seen in V2XSet \cite{xu2022v2xvit}, where sensors are clustered near the intersection center, and distributed placement, where sensors are spread out as shown in Fig. \ref{fig:sensor_placement}. 


The placement of infrastructure sensors, along with their associated parameters—including heading, relative positions among multiple sensors, and individual sensor configurations—significantly influences the system’s capability to accurately perceive road users \cite{vijay2021optimalplacement}. However, a full comprehensive evaluation of sensor placements across different intersections can be highly resource-intensive, such as requiring extensive data collection,  digital twin modeling, and iterative model training (used as benchmark in this study). Thus, an effective and scalable alternative evaluation method is critically needed.

This paper introduces a rapid \textbf{In}frastructrue \textbf{S}ensor \textbf{P}lacement \textbf{E}valuation (InSPE) framework for rapidly assessing perception effectiveness at safety-critical intersections. InSPE incorporates key metrics—including sensor coverage, occlusion analysis, and information gain—to comprehensively evaluate multi-modal infrastructure-based perception under different sensor placements. 
To support this, we developed a flexible data-generation tool that allows configurable sensor positions and parameters. Using this tool within the CARLA simulator \cite{dosovitskiy2017carla}, we created Infra-Set, a large-scale dataset covering 10 intersections with diverse geometries, traffic densities, and ambient environment and lighting conditions. 
Additionally, we conducted benchmarking experiments using state-of-the-art (SOTA) infrastructure-based multi-modal perception algorithms, evaluating the surrogate metrics across various sensor placements and configurations. 
The results demonstrate that our proposed surrogate metric enables fast and efficient analysis of sensor placements' perception capabilities at intersections. The contributions of this work are listed as follows: \begin{itemize}
    \item We propose a fast infrastructure sensor placement evaluation framework with a set of surrogate metrics, capable of rapidly analyzing and assessing perception capabilities for arbitrary intersection and sensor placement scenarios.
    \item We develop a flexible data generation tool and introduce Infra-Set, a large-scale dataset that comprehensively covers diverse intersection geometries, traffic scenarios, ambient environment and lighting conditions, and realistic sensor placements to advance infrastructure-based sensing and perception research.
    \item We employ heterogeneous, multi-modal perception algorithms over high-resolution intersection digital twins as benchmarks and conduct extensive experiments alongside quantitative analyses to rigorously validate our perception evaluation framework InSPE in its effectiveness in systematically evaluating the effects of various sensor placement strategies.
\end{itemize}

\section{Related Work}
\begin{figure*}[t]
\centering
\includegraphics[width=0.8\textwidth]{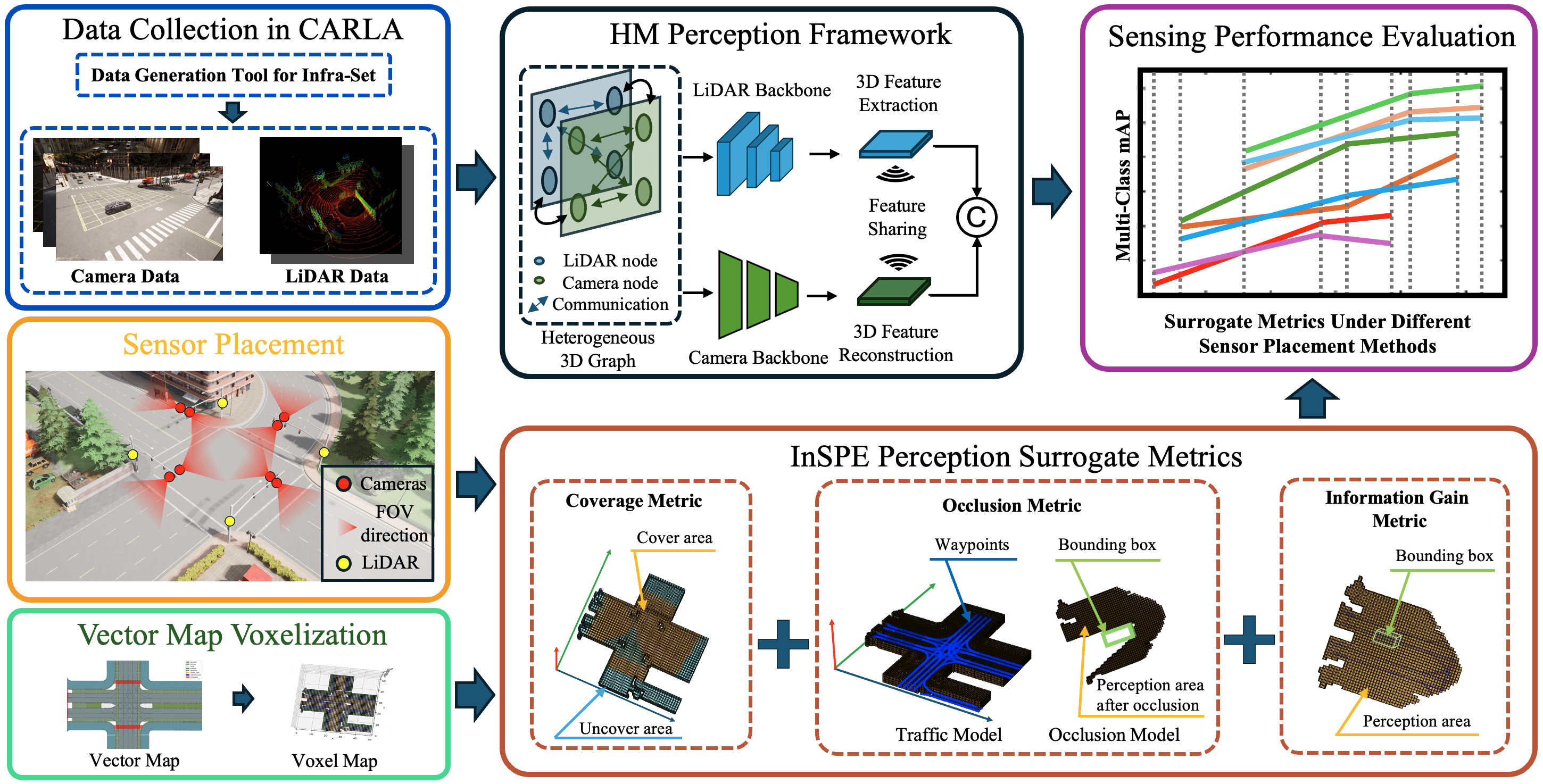}
\caption{{\bf Sensor Placement Evaluation Framework.} HM Perception framework refers to heterogeneous multi-model perception framework.}
\label{fig:workflow}
\vspace{-3mm}
\end{figure*}

\subsection{Sensor Placement for Perception Evaluation}  
Previous research on sensor placement has primarily focused on vehicle-mounted sensors.
Ma et al. ~\cite{ma2021perception} 
proposed a Bayesian theory-based conditional entropy approach to evaluate vehicle sensor placements. 
Other studies have utilized similar entropy-based approaches to analyze the role of multi-LiDAR~\cite{hu2021multi} and camera-LiDAR~\cite{li2024influence} sensor placements in vehicular perception.  For infrastructure-based perception, Kim et al. \cite{kim2023lidar} and SEIP~\cite{luo2023seip} adopted a voxel coverage method based on LiDAR sensors, whereas the work of Cai et al.~\cite{cai2023rls} employed an analysis of point cloud density distribution to assess the effectiveness of LiDAR placement. These studies only focus on LiDAR placement at junctions and do not evaluate sensor placement under heterogeneous and multi-modal conditions. To address these limitations, we introduce the InSPE surrogate metric in Section \ref{s_metrics}, designed for intelligent infrastructure. This metric quantitatively evaluates sensor placement beyond the junction area, capturing a broader perception of regions and providing a more comprehensive assessment framework.
 
\subsection{Datasets for Cooperative Perception}  
Existing datasets largely focus on vehicle-centric perception, offering limited support for infrastructure-based sensing. OPV2V~\cite{xu2021opv2v} is designed for Vehicle-to-Vehicle (V2V) perception and lacks infrastructure integration. 
Both simulation datasets—V2X-Set~\cite{xu2022v2xvit} and V2X-Sim~\cite{li2022v2xsim}—and real-world datasets—V2X-Real~\cite{xiang2024v2xreal}, V2X-PnP~\cite{zhou2024v2xpnp}, and DAIR-V2X~\cite{yu2022dairv2x}—support Vehicle-to-Everything (V2X) communication and include a single infrastructure sensor. However, they are limited in scale and are not specifically designed to support Infrastructure-to-Infrastructure (I2I) perception with flexible sensor placements.
R-Cooper ~\cite{hao2024rcooper} introduces road category classifications but lacks diverse intersection geometries, making it inadequate for large-scale heterogeneous sensor placement evaluation. To address these limitations, we introduce a large-scale I2I perception dataset built by our data generation tool. Unlike existing datasets, it enables dynamic sensor placement, supports multiple infrastructure units,  and provides a scalable benchmarking framework for infrastructure-based perception models.


\subsection{Cooperative Perception Algorithms}  
Cooperative perception fuses sensors data from multiple agents to extend detection ranges and mitigate occlusions. Existing methods such as OPV2V ~\cite{xu2021opv2v}, V2VNet ~\cite{wang2020v2vnet}, Where2comm ~\cite{hu2022where2comm} focus on LiDAR-based cooperative perception in connected vehicles. Meanwhile, multi-modal frameworks like BEVFusion ~\cite{liu2022bevfusion} and BEVFormer ~\cite{li2022bevformer} integrate LiDAR and camera data but are restricted to single-agent perception.  Several V2X fusion approaches, including V2X-ViT \cite{xu2022v2xvit}, DiscoNet \cite{li2022disconet}, CoAlign ~\cite{lu2023coalign}, Where2comm, and Who2comm, aim to enhance multi-agent cooperative perception by optimizing sensor fusion strategies. Compared to these vehicle-centric approaches, I2I perception allows sensors to be placed independently, enabling more cost-effective deployments by avoiding unnecessary LiDAR placement alongside every camera. While methods like HM-ViT \cite{hao2023hmvit} and HEAL \cite{lu2024eheal} explore heterogeneous multi-modal fusion, they do not address the unique challenges of I2I settings, such as flexible sensor distribution, vantage point selection, and coverage requirements. To bridge this gap, we inrtroduces a frameworks that supports benchmarking for  heterogeneous multi-modal perception in I2I settings.

\section{Method}

We propose a novel perception evaluation metric set specifically designed for diverse safety-critical intersections. The metric takes the basic vector map of the intersection and sensor placement and parameters as input and uses ray casting algorithms for sensing modeling.
The overall perception evaluation framework is illustrated in Figure~\ref{fig:workflow}.

\subsection{Problem Formulation}








To evaluate the perception performance of multi-modal sensor placement at intelligent intersections, we focus on detecting objects within ROI. The ROI is the intersection area with a radius $D_{inf}$ raging from 50 to 100 meters from the intersection center $(x_c,y_c)$, depending on intersection type and speed limits~\cite{sultana2014analysis,gorrini2016towards,easa2020reliability}. The area within 30 meters of the intersection is defined as the core region, which requires immediate safety awareness. We further define ROI as the 3D voxelized space $\Omega$ constrained by a 3D vector map, consisting of $N$ voxels $V_i$ each at location $(x_i, y_i, z_i) \in \mathbb{R}^3 $.
\begin{equation}
    \Omega = \left\{ \
    \begin{aligned}
    & \sqrt{(x_i - x_c)^2 + (y_i - y_c)^2} \leq D_{inf},\\
    & \text{ground} \leq z_i \leq \text{ground} + 4m, \\
    & \Omega = \left\{V_1,..,V_i,..V_N \right\}
    \end{aligned}
    \right\}.
\end{equation}

We define an Infrastructure Unit (IU) as:
\begin{equation}
\text{IU} = \left\{ s \in \mathcal{S} \ \middle|\ 
\begin{aligned}
&\forall\, s_i, s_j \in \text{IU}, \\
&\sqrt{(x_i - x_j)^2 + (y_i - y_j)^2} \leq 2m, \\
&|z_i - z_j| \leq 4m, \\
& p_i = p_j
\end{aligned}
\right\} , 
\end{equation}
where, $s$ represents a sensor, $i,j$ are different sensor indices, $\mathcal{S}$ is the set of all sensors in one IU, and $x,y,z$ represent the physical location of a sensor. The $p_i = p_j$ ensures that all sensors in the IU share the same processing unit.  

We formulate the heterogeneous multi-modal sensor placement problem as a 3D object detection performance challenge for multi-sensor fusion algorithms under a given intersection and its corresponding sensor placement. However, the diversity of real-world intersections, the complexity of sensor combinations, and the specificity of detection algorithms make large-scale real-world performance tests impractical. To overcome these challenges,  we propose a scalable surrogate metrics that simulates heterogeneous multi-infrastructure-unit, multi-modal sensor systems across various intersection types and benchmarks it against state-of-the-art (SOTA) perception algorithms.

\begin{figure}[t]
\centering
\begin{minipage}[b]{0.3\textwidth}
  \centering
  \includegraphics[width=\textwidth]{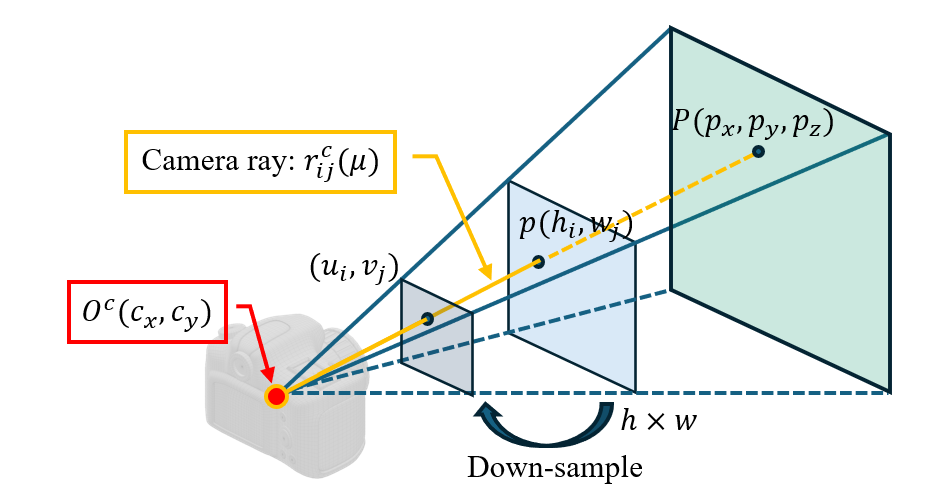}
  \subcaption{Camera View Frustum Model}
  \label{fig:sensing_a}
\end{minipage}
\hfill
\begin{minipage}[b]{0.48\textwidth}
  \centering
  \includegraphics[width=\textwidth]{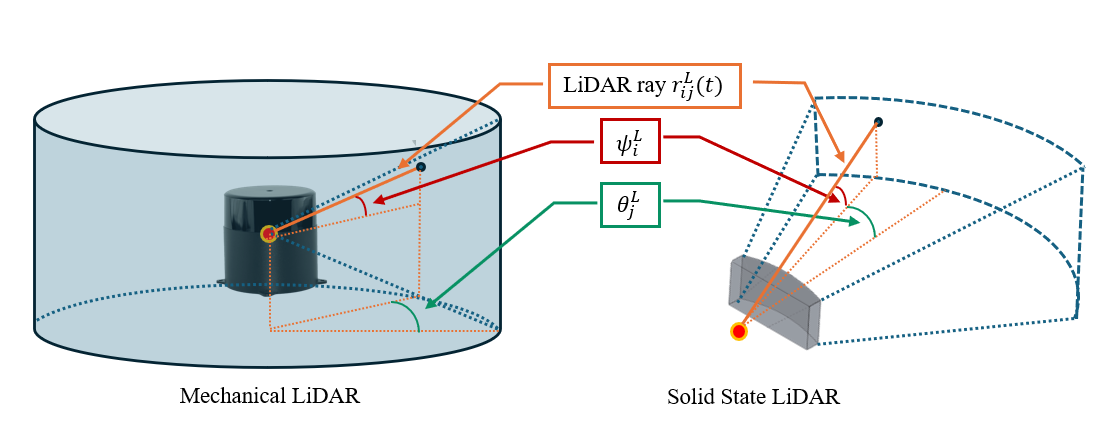}
  \subcaption{Mechanical and Solid State LiDAR Model}
  \label{fig:sensing_b}
\end{minipage}
\caption{\textbf{Illustration of Camera and LiDAR Sensing Model.} The red dot represents the center of the camera and LiDAR, and orange line is the camera and LiDAR ray.}
\label{fig:sensing}
\vspace{-4mm}
\end{figure}

\subsection{Camera and LiDAR Sensing Modeling}
\label{sec:cam_n_lidar}
To evaluate the sensing capability of multi-modal sensor placement at the intersection, we introduce a ray-cast sensing model for camera and LiDAR, building upon work 
\cite{hu2022investigating,li2024influence}. Furthermore, we employ the Bresenham algorithm \cite{bresenham1998algorithm} to solve the ray-cast model, identifying the set of voxels that lie within the sensor’s “view frustum”  or sensing range under given sensor placement $P_0$:
\begin{equation}
    \Omega|P_0 = \{V_1^{P_0},V_2^{P_0},...,V_n^{P_0}, n \in N\} .
\end{equation}

\noindent \textbf{Camera Sensing Model.} Based on the pinhole camera model \cite{hartley2003multiple}, we model the camera's field of view as a frustum with focal length $f$~\cite{kerlow2009art}. Furthermore, we formulate the camera perception model as a ray projection model composed of rays connecting the real-world coordinate $P(p_x, p_y, p_z)$, the camera pixel $p(h_i, w_j)$, and the camera principal point $\textbf{O}^C(c_x, c_y)$, as illustrated in the Figure~\ref{fig:sensing_a}. A ray-cast camera model is a geometric framework in which each pixel on the image plane is represented by a ray that originates at the camera center (or pinhole) and extends into the environment. Consequently, the entire camera field of view can be expressed as the collection of rays contained within the camera frustum.

For learning-based camera perception algorithms, an input image with the original resolution $h \times w$ is often resized to a smaller one, with the downsampling rate $\lambda$. Therefore, for each downsampled pixel $u_i, v_j$, where $i = 1, 2, \ldots, h\lambda$ and $j = 1, 2, \ldots, w\lambda$, we compute its corresponding ray as follows:
\begin{equation}
    \textbf{r}_{ij}^{C}(\mu) = \textbf{O}^C+\mu \cdot \textbf{d}_{ij} = \textbf{O}^C+\mu \cdot \begin{bmatrix}
\left(w_{j} - c_x\right)/f \\[6pt]
\left(h_{i} - c_y\right)/f \\[6pt]
1
\end{bmatrix}, 
\end{equation}
where, $\mu \geq$ 0, and $\textbf{d}_{ij}$ is direction vector.

\noindent \textbf{LiDAR Sensing Model.} Based on the construction of real rotating mechanical LiDARs and solid-state LiDARs, we model the LiDAR as a system composed of a large number of discrete beams. Each beam, indexed by $(i,j)$, can be regarded as a ray emanating from the sensor origin $\textbf{O}_L$, with its direction determined by the horizontal scanning angle $\theta_i^L$ and the vertical scanning angle $\psi_j^L$, as illustrated in the Figure~\ref{fig:sensing_b}. The LiDAR's sensing region can be described as a three-dimensional originating from vertex $O$, comprising all rays within a specified angular range, where $t_{\max}$ denotes the maximum sensing distance of the LiDAR. For a LiDAR with a horizontal scanning angle $\theta^L_0$ and a vertical scanning angle $\psi^L_0$, we compute the yaw $\theta^L_{j}$ and the pitch $\psi^L_{i}$ rotation angles for these rays as follows:
\begin{equation}
    \theta^L_{j} = -\frac{\theta^L_0}{2} + \frac{j\cdot\theta^L_0}{J},  \psi^L_{i} = -\frac{\psi^L_0}{2} + \frac{j\cdot\psi^L_0}{I},
\end{equation}
where, $j\in\{1,...,J\},i\in\{1,...,I\}$. Therefore, the parametric equation of the ray is: 
\begin{equation}
    r^{L}_{ij}(t) = \textbf{O}^L + t\cdot \textbf{d}_{ij} = \textbf{O}^L + t\begin{bmatrix}
\cos\psi^L_{i}\,\cos\theta^L_{j} \\
\cos\psi^L_{i}\,\sin\theta^L_{j} \\
\sin\psi^L_{i},
\end{bmatrix}, 
\end{equation}
where, $0 \le t \le t_{max}$.


\subsection{Perception Surrogate Metric Design}

\textbf{Perception Coverage Metric.}
In order to quantify the sensing coverage capability of sensors at an intersection, we use a perception coverage metric \( C \) for quantification. According to the sensor perception model defined in section \ref{sec:cam_n_lidar}, we represent the voxels that are traversed by the sensor rays as the visible region, while the regions that are not traversed are considered non-visible. Thus, we can define:
\[
f(V_i) = 
\begin{cases}
1, & \text{if sensor ray passes through \( V_i \), \( r_{ij} \in V_i \)}, \\
0, & \text{otherwise}.
\end{cases}
\]
Then, the sensor perception coverage can be defined as the proportion of all voxels that are effectively covered. Considering the varying importance or weight of different regions within the intersection, we introduce a weight function \( w(V_i) \). Therefore, the perception coverage is defined as the weighted coverage:
\begin{equation}
    \textit{C} = \frac{\sum_{V_i \in \Omega} w(V_i) \cdot f(V_i)}{\sum_{V_i \in \Omega} w(V_i)}.
\end{equation}
According to \cite{gattis2010guide,mcmahon2002analysis,kim2024enhancing}, we divide the intersection area into the following regions: driveway, junction, crosswalk, sidewalk, and shoulder. And based on the statistical analysis from \cite{kim2024enhancing, carter2006pedestrian}, we assign normalized weights based on their safety awareness level and importance as follows: $0.22 : 0.25 : 0.23 : 0.17 : 0.13$.


\noindent \textbf{Perception Occlusion Metric.}
Deploying multiple sensors at an intersection enhances perception by mitigating occlusion through diverse viewpoints. Therefore, we propose an intersection-based perception occlusion model and quantify it using the occlusion probability metric \( O \). Based on the individual sensor perception models, we first establish the occlusion interaction model between the sensor and the target.

\textit{ (a) Sensing Ray-Surface Intersection Model} 
\label{subsec:unified_formulation}

For both sensing modalities, let $\mathbf{O}$ denote the sensor origin and $\mathbf{d}$ represent the unit direction vector of an emitted ray. The parametric ray equation is expressed as:
\begin{equation}
    \mathbf{r}(t) = \mathbf{O} + t\mathbf{d}, \quad t \geq 0 ,
\end{equation}
where $t$ denotes the distance along the ray. The relationship between the occluder and the perceived ray reduces to solving:
\begin{equation}
    F\left(\mathbf{O} + t^*\mathbf{d}\right) = 0, \quad t^* \geq 0,
\label{eq:general_intersection}
\end{equation}
where $F(\mathbf{r})=0$ defines the implicit surface representation of objects, and $t^*$ corresponds to the first valid intersection distance. The collision position is then given by $\mathbf{r}(t^*)$.

\textit{(b) Waypoint-based Traffic Model} 

We establish a waypoint-based bounding box traffic model. This model utilizes the road waypoint information provided by the vector map to model vehicles in the intersection area as bounding boxes and employs waypoints to statistically describe the vehicles' positions, orientations, and trajectories. The specific method is as follows:
(a) Extract the waypoints \( w = (x_w, y_w, z_w) \) from the vector map that are located within the sensor's perception region, i.e., where \( f(V_i)=1 \).
(b) Since vehicles are the primary occlusion targets at intersections, we model the largest detectable vehicle using a bounding box, assuming its length, width, and height are \( L \), \( W \), and \( H \), respectively.
(c) Using the positional information from consecutive waypoints, the heading angle of the bounding box can be calculated as $\theta_w = f(w_t, w_{t+1})$.
(d) Based on the continuous waypoint information, we can model the traffic flow of vehicle bounding boxes within the perception region.

\textit{(c) Occlusion Probability Calculation}

By combining the sensing ray-surface intersection model and the traffic model, we can compute the occluded perception region. Let \( V_{\text{orig}}^{(k)} \) denote the set of voxels corresponding to the original (unoccluded) target area in the \( k \)-th continuous waypoint frame, and let \( V_{\text{occ}}^{(k)} \) denote the set of voxels that are occluded in that frame. Then, the occlusion ratio for the \( k \)-th waypoint frame is defined as:
\begin{equation}
    O^{(k)} = \frac{|V_{\text{occ}}^{(k)}|}{|V_{\text{orig}}^{(k)}|},
\end{equation}
where \(|\cdot|\) denotes the area (or count) of voxels. To capture the effect of different sensor positions over the entire ROI, we average the occlusion ratios over \( N \) continuous waypoint frames:
\begin{equation}
O = \frac{1}{N} \sum_{k=1}^{N} 1- O^{(k)} = \frac{1}{N} \sum_{k=1}^{N} 1-\frac{|V_{\text{occ}}^{(k)}|}{|V_{\text{orig}}^{(k)}|}.
\end{equation}



    
    



\noindent \textbf{Information Gain Metric.}
To evaluate the efficacy of the sensor placement in reducing perceptual uncertainty, based on previous work \cite{ma2021perception,hu2022investigating, li2024influence} and introduce a modified information gain (IG) metric. The metric intergrates the voxel map with a waypoint-based traffic model to quantify the reduction in uncertainty regarding the occupancy state of ROI when sensor observations are incorporated. Based on the traffic model, the occupancy probability for each voxel \(V_i\) is estimated from the statistical distribution of vehicle bounding boxes moving along the predefined waypoints over multiple frames. Let \(T\) denote the number of frames and define the time-averaged occupancy probability for voxel \(V_i\) as
\begin{equation}
    \hat{p}(V_i) = \frac{1}{T} \sum_{t=1}^{T} \mathbf{1}\bigl( V_i^{(t)} \text{ is occupied} \bigr),
\end{equation}
where \(\mathbf{1}(\cdot)\) is the indicator function and we denote $\hat{p}(V_i)$ as $\hat{p}$ . Then, the total entropy of the voxel map, incorporating temporal information, is computed as:
\begin{equation}
    H(\Omega) = \sum_{i=1}^{N} H(V_i) = -\sum_{i=1}^{N} \left[ \hat{p} \log \hat{p} + \left(1-\hat{p}\right) \log \left(1-\hat{p}\right) \right].
\end{equation}

Given a sensor placement $P_0$, the aggregated conditional entropy within the sensor’s perception region is then computed as:
\begin{equation}
    H(\Omega|P_0) = \sum_{V_i^{P_0} \in \Omega|P_0} H(V_i^{P_0}),
\end{equation}
where $ V_i^{P_0}$ denotes each voxel in the sensor's field of view.


The information gain due to the sensor placement \( P_0 \) is defined as the difference between the total entropy of the occupancy grid and the conditional entropy given the sensor’s observations:
\begin{equation}
    IG_{\Omega, P_0} = H(\Omega) - H(\Omega|P_0).
\end{equation}
Since \( H(\Omega) \) is invariant to the sensor placement, \( IG_{\Omega, P_0} \) quantifies the reduction in uncertainty provided solely by the sensor’s perception. To facilitate comparisons across different surrogate metrics, the information gain is further normalized to yield a metric bounded within \([0,1]\):
\begin{equation}
    IG = \frac{IG_{\Omega, P_0}}{H(\Omega)} = 1 - \frac{H(\Omega|P_0)}{H(\Omega)}.
\end{equation}
In this formulation, an \( IG_{\text{norm}} \) value of 1 indicates a complete reduction in uncertainty (i.e., the sensor placement fully resolves the occupancy state), whereas a value of 0 implies no reduction in uncertainty.

\subsection{Surrogate Metrics} \label{s_metrics}
To balance the varied impacts and representations of the aforementioned surrogate metrics on sensing performance, we adopt a weighted fusion method to compute the final surrogate metric. Specifically, each surrogate metric characterizes the sensor's performance in a specific aspect (e.g., sensing coverage, occlusion, detection information uncertainty, etc.), and these metrics exhibit different degrees of importance and sensitivity in practical applications. To ensure that the aggregated metric comprehensively and accurately reflects the overall sensing capability of the sensor, the weight of each metric can be dynamically adjusted based on the traffic conditions of the scenario and statistical analysis results. We compute the perception surrogate metric as shown below:
\begin{equation}
    P_{sm} = w_c \cdot C + w_o \cdot O + w_{ig} \cdot IG,
\end{equation}
where, $w_c + w_o + w_{ig} = 1$. In our work, the recommended weights are: $w_c:w_o:w_{ig} = 0.3:0.5:0.2$.

\section{Experiments}

\begin{table*}[t]
\centering
\caption{\textbf{Quantitative 3D Detection Results on Infra-Set}. The table shows the NuScenes mAP(\%) results. "S.P." represents sensor placement. * in the table means that the algorithm was modified to adapt to Camera+LIDAR heterogeneous multi-modal settings.}
\label{table1}
\resizebox{\linewidth}{!}{
\begin{tabular}{|c||ccc|ccc|ccc|ccc|}
\hline
mAP (\%)     & \multicolumn{3}{c|}{Car}         & \multicolumn{3}{c|}{Pedestrian} & \multicolumn{3}{c|}{Cyclist}   & \multicolumn{3}{c|}{Truck} \\ \hline
\diagbox{Model}{S.P.} & \textbf{Cam-c} & \textbf{Cam-d1} & \textbf{Cam-d2} 
& \textbf{Cam-c} & \textbf{Cam-d1} & \textbf{Cam-d2} 
& \textbf{Cam-c} & \textbf{Cam-d1} & \textbf{Cam-d2} 
& \textbf{Cam-c} & \textbf{Cam-d1} & \textbf{Cam-d2} \\ \hline

LSS-Eff \cite{tan2019efficientnet} & \multicolumn{1}{c|}{31.88} & \multicolumn{1}{c|}{45.70} & \multicolumn{1}{c|}{37.16} & \multicolumn{1}{c|}{22.83} & \multicolumn{1}{c|}{11.60} & \multicolumn{1}{c|}{16.01} & \multicolumn{1}{c|}{10.76} & \multicolumn{1}{c|}{17.17} & \multicolumn{1}{c|}{40.88} & \multicolumn{1}{c|}{22.06} & \multicolumn{1}{c|}{42.79} & \multicolumn{1}{c|}{42.86} \\ \hline
LSS-ResNet \cite{he2016deep} & \multicolumn{1}{c|}{19.09} & \multicolumn{1}{c|}{52.36} & \multicolumn{1}{c|}{43.53} & \multicolumn{1}{c|}{3.95} & \multicolumn{1}{c|}{27.84} & \multicolumn{1}{c|}{29.64} & \multicolumn{1}{c|}{1.77} & \multicolumn{1}{c|}{17.74} & \multicolumn{1}{c|}{25.00} & \multicolumn{1}{c|}{19.32} & \multicolumn{1}{c|}{53.57} & \multicolumn{1}{c|}{52.97} \\ 
\hline\hline
\diagbox{Model}{S.P.} & 
\textbf{L-c} & \textbf{L-d1} & \textbf{L-d2} & 
\textbf{L-c} & \textbf{L-d1} & \textbf{L-d2} & 
\textbf{L-c} & \textbf{L-d1} & \textbf{L-d2} & 
\textbf{L-c} & \textbf{L-d1} & \textbf{L-d2} \\ \hline

V2VNet \cite{wang2020v2vnet} & \multicolumn{1}{c|}{63.87} & \multicolumn{1}{c|}{82.60} & \multicolumn{1}{c|}{71.62} & \multicolumn{1}{c|}{49.81} & \multicolumn{1}{c|}{74.09} & \multicolumn{1}{c|}{66.73} & \multicolumn{1}{c|}{13.92} & \multicolumn{1}{c|}{28.94} & \multicolumn{1}{c|}{24.61} & \multicolumn{1}{c|}{45.79} & \multicolumn{1}{c|}{53.99} & \multicolumn{1}{c|}{48.60} \\ \hline
V2X-ViT \cite{xu2022v2xvit} & \multicolumn{1}{c|}{69.62} & \multicolumn{1}{c|}{85.50} & \multicolumn{1}{c|}{75.43} & \multicolumn{1}{c|}{43.48} & \multicolumn{1}{c|}{75.85} & \multicolumn{1}{c|}{52.68} & \multicolumn{1}{c|}{24.16} & \multicolumn{1}{c|}{48.10} & \multicolumn{1}{c|}{29.79} & \multicolumn{1}{c|}{47.55} & \multicolumn{1}{c|}{58.73} & \multicolumn{1}{c|}{47.98} \\ \hline
CoAlign \cite{lu2023coalign} & \multicolumn{1}{c|}{71.68} & \multicolumn{1}{c|}{86.17} & \multicolumn{1}{c|}{87.92} & \multicolumn{1}{c|}{50.05} & \multicolumn{1}{c|}{77.51} & \multicolumn{1}{c|}{70.55} & \multicolumn{1}{c|}{21.31} & \multicolumn{1}{c|}{55.03} & \multicolumn{1}{c|}{46.28} & \multicolumn{1}{c|}{47.05} & \multicolumn{1}{c|}{63.00} & \multicolumn{1}{c|}{55.64} \\ 
\hline\hline
\diagbox{Model}{S.P.} & 
\textbf{Cam-c/L-c} & \textbf{Cam-d1/L-d1} & \textbf{Cam-d2/L-d2} & 
\textbf{Cam-c/L-c} & \textbf{Cam-d1/L-d1} & \textbf{Cam-d2/L-d2} & 
\textbf{Cam-c/L-c} & \textbf{Cam-d1/L-d1} & \textbf{Cam-d2/L-d2} & 
\textbf{Cam-c/L-c} & \textbf{Cam-d1/L-d1} & \textbf{Cam-d2/L-d2} \\ \hline

AttFuse$^{*}$ \cite{xu2021opv2v} & \multicolumn{1}{c|}{65.56} & \multicolumn{1}{c|}{86.33} & \multicolumn{1}{c|}{83.19} & \multicolumn{1}{c|}{52.02} & \multicolumn{1}{c|}{75.68} & \multicolumn{1}{c|}{72.72} & \multicolumn{1}{c|}{29.66} & \multicolumn{1}{c|}{54.60} & \multicolumn{1}{c|}{51.34} & \multicolumn{1}{c|}{52.94} & \multicolumn{1}{c|}{64.92} & \multicolumn{1}{c|}{61.59} \\ \hline
DiscoNet$^{*}$ \cite{yuan2022keypoints} & \multicolumn{1}{c|}{69.98} & \multicolumn{1}{c|}{90.18} & 87.29 & \multicolumn{1}{c|}{46.00} & \multicolumn{1}{c|}{76.84} & 73.96 & \multicolumn{1}{c|}{25.79} & \multicolumn{1}{c|}{60.82} & 58.31 & \multicolumn{1}{c|}{51.80} & \multicolumn{1}{c|}{58.86} & 57.57 \\ \hline
HM-ViT$^{*}$ \cite{hao2023hmvit} & \multicolumn{1}{c|}{72.56} & \multicolumn{1}{c|}{90.61} & 88.69 & \multicolumn{1}{c|}{51.26} & \multicolumn{1}{c|}{78.98} & 76.76 & \multicolumn{1}{c|}{30.99} & \multicolumn{1}{c|}{65.51} & 58.93 & \multicolumn{1}{c|}{58.43} & \multicolumn{1}{c|}{70.19} & 66.26 \\ \hline
\end{tabular}
}
\end{table*}

\subsection{Infra-Set: An open dataset for infrastructure-based multi-modality research}

To the best of our knowledge, there exists no public smart intersection-based datasets suitable for studying heterogeneous, multiple IUs, multi-modal sensor placement research. To better verify our proposed method, we designed an automatic multi-modal sensor placement data collection tool based on the high-fidelity CARLA simulation environments and digital twins. With the tool and simulator, we build our dataset, Infra-Set, following the V2XSet \cite{xu2022v2xvit} data format. We collected data for three distinct traffic flow scenarios—low, medium, and high density—from 10 different intersections within CARLA Towns 3, 4, 5, 6, 7, and 10.
The average traffic density for each scenario comprised approximately 40 objects, including various categories such as pedestrians, vehicles, trucks, and cyclists. Each scenario encompassed three different lighting conditions: midday, nighttime, and dusk. The sensors were positioned on traffic lights, utility poles, or appropriate structures. Infra-Set consists of approximately 144,000 scene frames. We split train:valid:test set into 7:1:2. For more details and visualization of our dataset, please refer to the 
supplementary material.
\subsection{Experiment Setup}
\textbf{Sensor Placement.} 
Inspired by the sensor distribution patterns observed in various datasets, we have designed three distinct camera placement strategies, as illustrated in Fig.\ref{fig:sensor_placement}. These three strategies include centralized camera placement (Cam-c), distributed camera placement 1 (Cam-d1), and distributed camera placement 2 (Cam-d2). Specifically, the centralized method aims to arrange the cameras in a manner similar to an in-vehicle camera system, concentrating them as much as possible while positioning the IU near the geometric center of the intersection. Distributed camera placement 1 distributes the cameras across traffic signal poles at the intersection, ensuring that their fields of view (FoV) are aligned with the primary traffic flow direction. Distributed 2 situates the cameras on roadside utility poles, strategically covering key areas of interest such as crosswalks, junctions, and driveways to enhance monitoring efficiency. Similarly, for LiDAR placement, we have devised three different installation schemes: centralized LiDAR configuration (L-c), decentralized LiDAR configuration 1 (L-d1), and decentralized LiDAR configuration 2 (L-d2).

Furthermore, for the combined camera-LiDAR configurations, we have integrated the three basic schemes to form three composite configurations: centralized camera and LiDAR (Cam-c/L-c), distributed camera 1 and distributed LiDAR 1 (Cam-d1/L-d1), and distributed camera 2 and distributed LiDAR 2 (Cam-d2/L-d2). For four-way or five-way intersections, the total number of cameras is maintained at 8 for each configuration, while the LiDAR setups consist of either one 64-beam LiDAR, four 32-beam LiDARs, or two 64-beam LiDARs, respectively. In the case of T-intersections, triangular intersections, or smaller four-way intersections, the number of cameras is correspondingly reduced, with the total number adjusted to 6. 



\begin{table*}[t]
\caption{\textbf{Perception Surrogate Metrics Under Different Sensor Placement Methods.} $C$ is the coverage metric, $O$ means the occlusion metric, $IG$ represents the information gain metric, and $P_{sm}$ is the perception surrogate metric. The numerical results in the table contain both mean $\pm$ standard deviation. }
\label{table2}
\centering
\resizebox{0.8\linewidth}{!}{
\begin{tabular}{@{}|c||c|c|c|c|c|c|c|c|c|@{}}
\toprule
\begin{tabular}[c]{@{}c@{}}Surrogate \\ Metrics\end{tabular} & Cam-c & Cam-d1 & Cam-d2 & L-c   & L-d1  & L-d2  & \begin{tabular}[c]{@{}c@{}}Cam-c\\ L-c\end{tabular} & \begin{tabular}[c]{@{}c@{}}Cam-d1\\ L-d1\end{tabular} & \begin{tabular}[c]{@{}c@{}}Cam-d2\\ L-d2\end{tabular} \\ \midrule
$C$                                                            &  0.605 $\pm$ \text{\scriptsize 0.025}    &   0.680 $\pm$ \text{\scriptsize 0.069}     &   0.691 $\pm$ \text{\scriptsize 0.046}     & 0.626 $\pm$ \text{\scriptsize 0.018} & 0.872 $\pm$ \text{\scriptsize 0.026} & 0.794 $\pm$ \text{\scriptsize 0.020} &           0.815 $\pm$ \text{\scriptsize 0.028}                                          &                        0.917 $\pm$ \text{\scriptsize 0.025}                               &          0.899 $\pm$ \text{\scriptsize 0.027}                                             \\ \midrule
$O$    &   0.502  $\pm$ \text{\scriptsize 0.017}  &   0.549  $\pm$ \text{\scriptsize 0.033}   &   0.514  $\pm$ \text{\scriptsize 0.034}   &    0.609 $\pm$ \text{\scriptsize 0.017}  &  0.794  $\pm$ \text{\scriptsize 0.040}   &   0.721  $\pm$ \text{\scriptsize 0.032}  &     0.703  $\pm$ \text{\scriptsize 0.015}          &     0.901 $\pm$ \text{\scriptsize 0.032}                                                  &   0.870 $\pm$ \text{\scriptsize 0.036}                                                    \\ \midrule
$IG$                                                           &   0.253 $\pm$ \text{\scriptsize 0.019}    &    0.295 $\pm$ \text{\scriptsize 0.025}    &     0.306 $\pm$ \text{\scriptsize 0.038}   & 0.442 $\pm$ \text{\scriptsize 0.012} & 0.679 $\pm$ \text{\scriptsize 0.022} & 0.555 $\pm$ \text{\scriptsize 0.010 }&    0.509 $\pm$ \text{\scriptsize 0.010 }                                                 &                0.724 $\pm$ \text{\scriptsize 0.023 }                                       &          0.637 $\pm$ \text{\scriptsize 0.030 }                                             \\ \midrule
$P_{sm}$       &   0.483 $\pm$ \text{\scriptsize 0.020}  &   0.538 $\pm$ \text{\scriptsize 0.042}    &  0.526 $\pm$ \text{\scriptsize 0.038}    &  0.581 $\pm$ \text{\scriptsize 0.16}   &  0.794 $\pm$ \text{\scriptsize 0.032}   &  0.710  $\pm$ \text{\scriptsize 0.024}  &       0.698    $\pm$    \text{\scriptsize 0.018}                                      &                            0.871    $\pm$         \text{\scriptsize 0.028}              &                                     0.832  $\pm$   \text{\scriptsize 0.032}             \\ \bottomrule
\end{tabular}
}
\end{table*}

\begin{figure*}[t]
\centering
\includegraphics[width=0.8\textwidth]{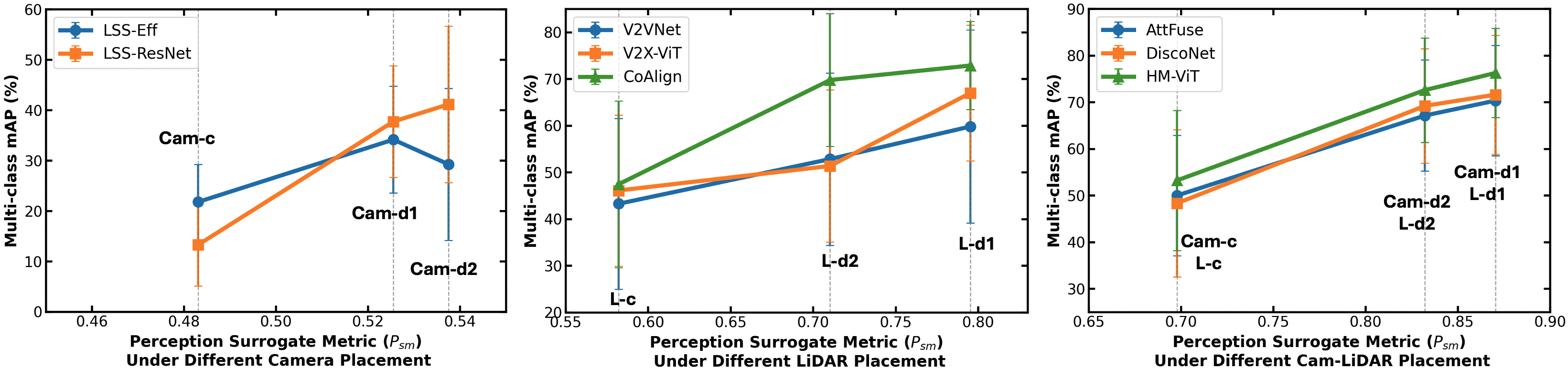}
\vspace{-2mm}
\caption{The relationship between perception surrogate metrics and multi-class mAP under different sensor placements.}
\label{fig:relationship}
\vspace{-3mm}
\end{figure*}

\noindent \textbf{Benchmarking Algorithms.}
In order to ensure a fair comparison of the 3D detection and sensing capabilities across different sensor placements at smart intersections, we conducted a comprehensive benchmark using camera-LiDAR detection algorithms specifically designed for heterogeneous, multi-agent, and multi-modal scenarios. For the camera-based detection pipeline, we employed two different backbone networks: Lift-Splat-Shoot (LSS) \cite{philion2020lift} with EfficientNet \cite{tan2019efficientnet} and LSS with ResNet101 \cite{he2016deep}. 
For sensor placements that rely solely on LiDAR, our fusion strategy incorporates state-of-the-art methods, such as V2VNet \cite{wang2020v2vnet}, V2X-ViT \cite{xu2022v2xvit}, and CoAlign \cite{lu2023coalign}. When evaluating the camera-LiDAR combination, we adopted classical feature fusion algorithms, including AttFuse \cite{xu2021opv2v}, DiscoNet \cite{yuan2022keypoints}, and HM-ViT \cite{hao2023hmvit}, which are well-regarded for their ability to effectively integrate heterogeneous sensor data. This systematic benchmarking approach enables a fair comparison of the performance trade-offs associated with each sensor placement and fusion strategy.

\noindent \textbf{Detection Performance Metrics.} Since our dataset comprises not only cars but also other detection targets such as trucks, pedestrians, and cyclists, and encompasses traffic scenarios with varying density levels, we employ the mean average precision (mAP) metric from the nuScenes benchmark\cite{caesar2020nuscenes}, which is specifically designed for 3D detection in autonomous driving. Unlike traditional intersection over union (IoU)-based metrics, which may fall short in fully capturing the nuances of detection performance in complex environments, the nuScenes mAP provides a more comprehensive evaluation of the detection algorithms' performance under a diverse range of challenging conditions.

\subsection{Quantitative Evaluation and Analysis.} 

\noindent \textbf{Main performance evaluation on different sensor placements.} In Table \ref{table1}, we present the 3D target detection results for nine different sensor placement configurations at various intersections. We observe that detection performance varies significantly with different sensor placements. Specifically, when LiDAR is included in the sensor combination, the detection performance for multiple classes improves by 20 $\sim$ 40\% compared to using cameras alone. Based on our sensor perception model analysis, this may be because current cameras have an effective perception depth of approximately 10$\sim$30 meters under varying lighting conditions, and their detection performance deteriorates markedly with increasing distance. In contrast, a single 32-beam LiDAR can cover up to 50 meters, and a 64-beam LiDAR can cover up to 100 meters. Therefore, even an infrastructure deployment consisting solely of multiple cameras does not match the sensing capability of a configuration that includes LiDAR. Moreover, the algorithm transforms 2D features into a 3D spatial representation, adding further complexity. We also note that although adding cameras to a LiDAR-based configuration improves detection performance, the gain is not as pronounced as that achieved when transitioning from a camera-only to a LiDAR-inclusive configuration. This is consistent with our previous work in HM-ViT \cite{hao2023hmvit}, which proves the dominating nature of LiDAR in improving mAP. However, this paper only uses mAP to validate the effectiveness of InSPE metrics by understanding the correlation. It is worth noting that additional metrics, such as detection of the existence of an object, where cameras may perform better in certains scenarios (e.g., far-away lcations where LiDAR point clouds can be sparse).

Furthermore, we observe that even with the same sensor type, detection results differ with various placement combinations. For cameras, distributed placements consistently yield better performance than centralized ones; however, for cam-d1, where the cameras face directly toward the road, vehicle detection is enhanced, whereas for cam-d2, where cameras cover more of the crosswalk and shoulder areas, the detection of small targets is improved. Regarding LiDAR, it is evident that increasing the number of LiDAR sensors leads to significant performance gains. However, even though the LiDAR in the L-c configuration uses a higher-grade 64-beam LiDAR with a 100-meter range, its performance is still inferior to that of the L-d2 configuration, which employs two lower-grade 32-beam LiDARs. Moreover, we find that, in the intersections of the InfraSet dataset, the performance improvement achieved by using four LiDARs in the L-d1 configuration is not as significant as that from L-c to L-d2. This further indicates that increasing the number of LiDAR sensors at an intersection is not necessarily beneficial; rather, sensor placement should be arranged appropriately based on factors such as intersection size, geometric layout, and type.

\noindent \textbf{Correlation between surrogate metric and perception performance.}
We analyze the relationship between detection performance and perception surrogate metrics under different sensor placements. Table \ref{table2} comprehensively presents the calculated perception surrogate metrics based on various sensor placement combinations. Fig. \ref{fig:relationship} illustrates the multi-class detection performance across different surrogate metrics and sensor configurations. It is evident that, although there are some fluctuations in detection performance, the overall positive correlation between surrogate metrics and detection performance is quite pronounced. These fluctuations, aside from arising from the randomness in dataset sampling and the uncertainties in the model training process, are primarily attributed to the variations in sensor placement at each intersection. Due to the differing geometric shapes and sizes of intersections, the positions of our infrastructure sensors vary, particularly for cameras, whose angles relative to the road surface are not always consistent. For learning-based perception algorithms, specific sensor placements can complicate the learning process. Consequently, our surrogate metrics reflect this to some extent in the statistical results. As shown in Table \ref{table2}, the standard deviation for camera-only placements is greater than that for LiDAR-only placements, as our LiDARs maintained consistent rotation parameters during placement, with differences mainly stemming from infrastructure locations and intersection characteristics. 

\section{Conclusions}

In this paper, we investigate the heterogeneous multi-model sensor placement problem at intelligent intersections. We propose a novel sensor placement evaluation framework specifically designed for intelligent infrastructure to assess the perception capabilities under various sensor placement situations at different intersections. To validate our approach, we developed a data generation tool capable of producing large-scale infrastructure-centric datasets suitable for diverse sensor placement methods. Extensive experiments were conducted using modified heterogeneous multi-modal benchmarking algorithms to examine the relationship between perception surrogate metrics and 3D perception capacity. Our experiments provide new insights and directions for future research on sensor placement at intersections.

\clearpage
\setcounter{page}{1}
\maketitlesupplementary

\section{Infra-Set Dataset Details}
\label{sec:infra-set-dataset}
\subsection{Intersection Selection}

\begin{figure}[h]
\centering
\includegraphics[width=0.48\textwidth]{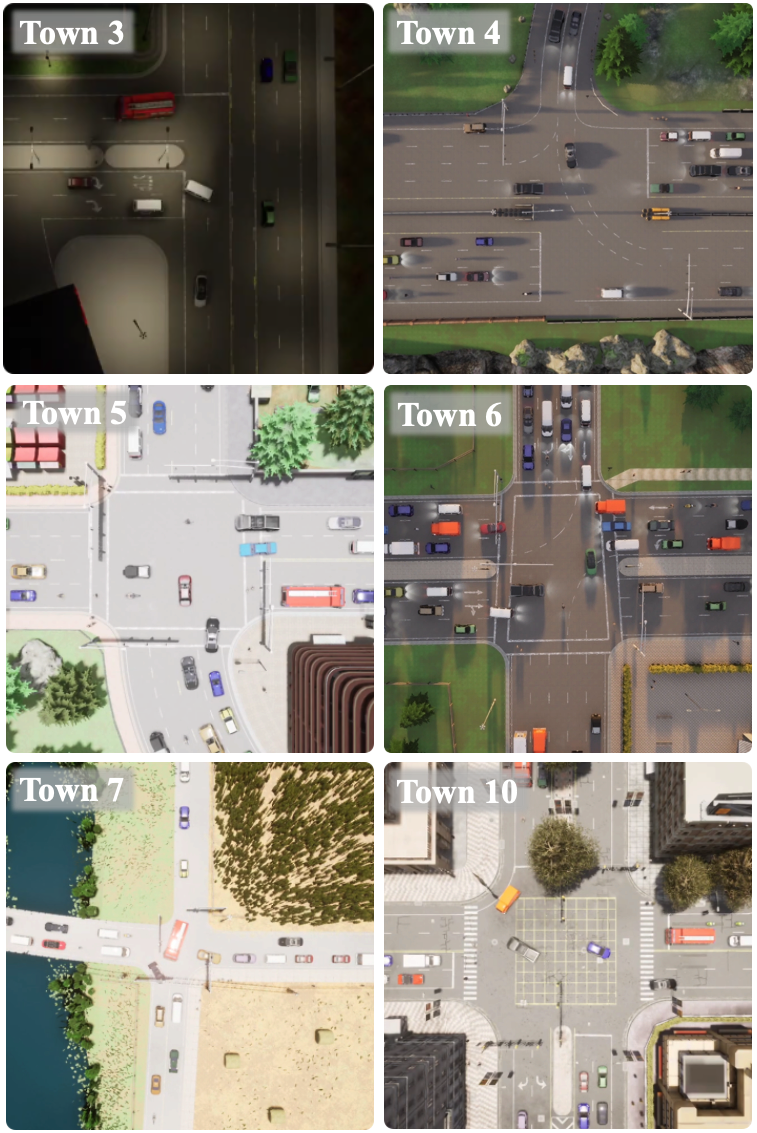}
\caption{{\bf Intersections and Their Traffic Flow in Different CARLA Towns.} The images feature intersections with various geometries, traffic flow conditions, and lighting conditions from different CARLA towns. In the Inf-Set dataset, we include various categories, such as pedestrians, vehicles, trucks, and cyclists. }
\label{fig:dataset}
\vspace{-3mm}
\end{figure}

To ensure diversity in the selected intersections for our dataset, we carefully chose intersections from CARLA towns 3, 4, 5, 6, 7, and 10. The visualizations of some selected intersections are shown in Fig. 
\ref{fig:dataset}. Among the selected 10 towns, the dataset includes four 4-way intersections, two T-intersections, one bridge entry intersection, one roundabout, one 5-way intersection, and one highway entry T-intersection.  

Furthermore, these 10 intersections were categorized by area into four large intersections, four medium-sized intersections, and two small intersections. In terms of environmental classification, the dataset comprises six urban intersections, three highway intersections, and one rural intersection.

\subsection{Data Size}
Overall, the Infra-Set dataset comprises 144,000 scenario frames. Each scenario contains camera or LiDAR data generated from at least nine different sensor placement methods (e.g., Cam-c, Cam-d1, Cam-d2, Cam-d3, L-c, L-d1, L-d2, etc.). The total data volume reaches 2.6 TB.

\begin{figure}[h]
    \centering
    \includegraphics[width=0.48\textwidth]{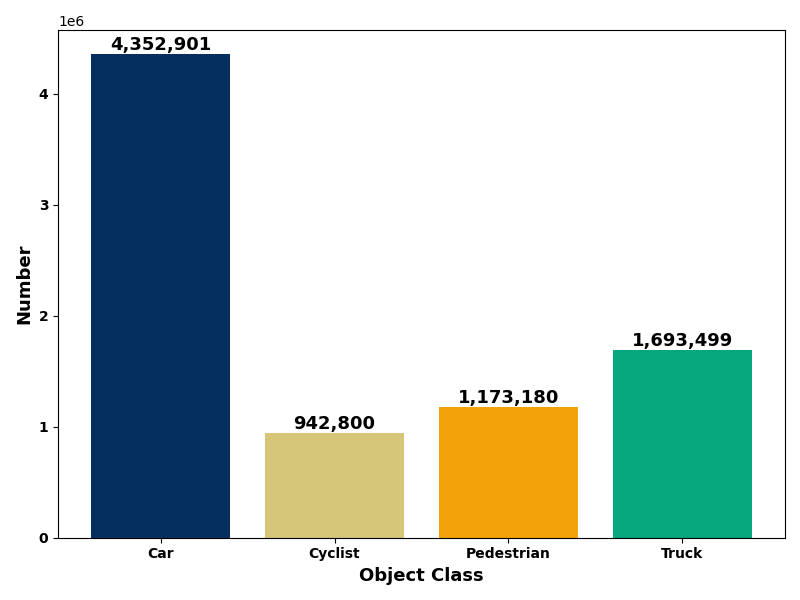}
    \caption{\textbf{Distribution of Total Object Counts Across Different Classes.} This figure illustrates the frequency distribution of objects across various categories, including cars, pedestrians, cyclists, and trucks, within the dataset.}
    \label{fig:count}
    \vspace{-3mm}
\end{figure}

\begin{figure*}[h]
    \centering
    \includegraphics[width=1\textwidth]{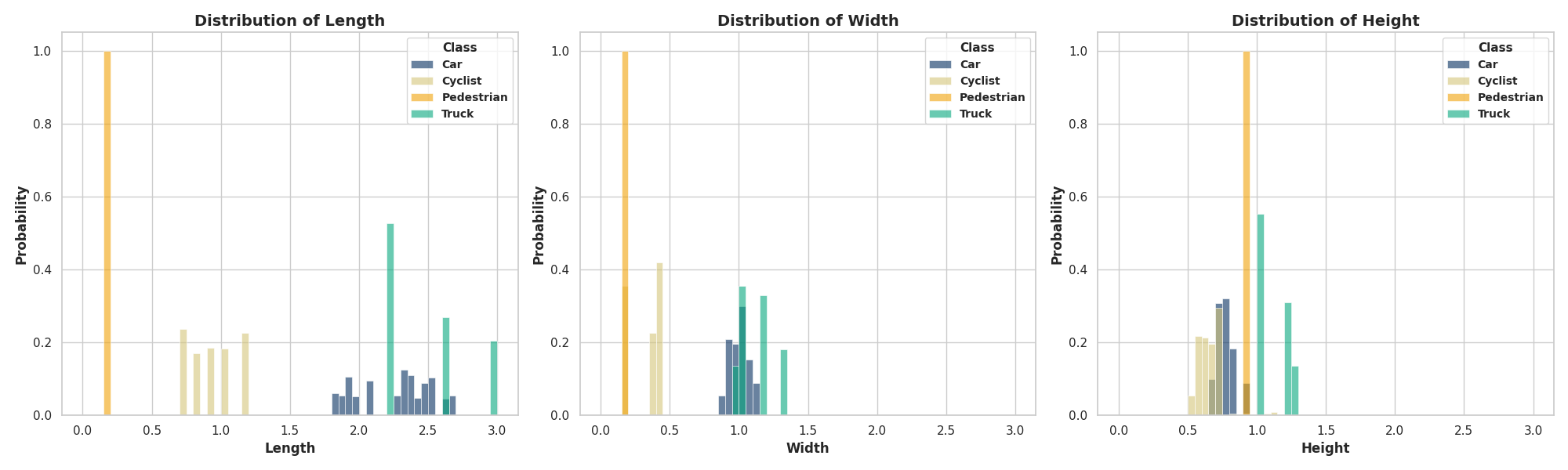}
    \caption{\textbf{Distribution of Object Dimensions Across Different Classes.} The figure presents the statistical distribution of object dimensions (length, width, and height) for different object classes. This provides insight into the physical characteristics of the objects in the dataset.}
    \label{fig:distribution}
    \vspace{-3mm}
\end{figure*}

\begin{table*}[t]
    \centering
    \renewcommand{\arraystretch}{1.1}
    \setlength{\tabcolsep}{3pt}
    \begin{tabular}{cccccccccc}
        \toprule
        \textbf{Dataset type} & \textbf{Source} & \textbf{Dataset} & \textbf{Year} & \textbf{Coop Mode} & \textbf{RGBs} & \textbf{LiDARs} & \textbf{Infra} & \textbf{Intersection} & \textbf{Det task} \\
        \midrule
        \multirow{6}{*}{Cooperative} 
            & \multirow{3}{*}{sim} & OPV2V \cite{xu2021opv2v} & 2022 & V2V & 44k & 11k & 4 & - & 3D \\
            & & V2X-Sim \cite{li2022v2xsim} & 2022 & V2X & 60k & 10k & 1 & 6 & 3D \\
            & & V2XSet \cite{xu2022v2xvit} & 2022 & V2X & 44k & 11k & 3 & 6 & 3D \\
        \cmidrule{2-10}
            & \multirow{2}{*}{real} & DAIR-V2X \cite{yu2022dairv2x} & 2022 & V2X & 39k & 39k & 4 & 28 & 3D \\
            & & V2V4Real \cite{xu2023v2v4real} & 2023 & V2V & 40k & 20k & 2 & - & 3D \\
            & & V2X-Real \cite{xiang2024v2xreal} & 2024 & V2X & 171k & 33k & 2 & 1 & 3D \\
        \midrule
        \multirow{2}{*}{Infra-Based} 
            & real & Rcooper \cite{hao2024rcooper} & 2024 & Infra & 50k & 30k & \textbf{2/4} & 1 & 3D \\
            & sim & \textbf{Infra-Set (Ours)} & \textbf{2025} & \textbf{Infra} & \textbf{3,546k} & \textbf{1,008k} & \textbf{2$\sim$8} & \textbf{10} & \textbf{3D} \\
        \bottomrule
    \end{tabular}
    \caption{\textbf{Comparisons of Representative Public Cooperative Perception Datasets for Autonomous Driving.} This table compares various cooperative perception datasets, categorizing them by dataset type (cooperative or infrastructure-based), source (simulation or real-world), and key properties such as year, cooperation mode, sensor availability, infrastructure support, and detection task type. Infra stands for Infrastructure.}
    \label{tab:dataset_comparison}
\end{table*}
\subsection{Data Anlysis}
Our dataset comprises three distinct traffic flow densities: high, medium, and low. In high-density scenarios, each scene contains an average of approximately 60 objects; in medium-density scenarios, around 40 objects; and in low-density scenarios, about 20 objects on average \cite{mcghee1998traffic, li2024multimodal}. The dataset mainly includes four object categories: car, pedestrian, cyclist, and truck, with the number of ground truth instances for each category shown in Fig. \ref{fig:count}.  The proportion of our dataset is constructed following the distribution of other established cooperative perception datasets \cite{xu2021opv2v, li2022v2xsim, xu2022v2xvit, yu2022dairv2x, xu2023v2v4real, xiang2024v2xreal}, ensuring a balanced representation of real-world autonomous driving scenarios across different environments and intersection types. Fig. \ref{fig:distribution} depicts the distribution of bounding box sizes. The figure illustrates the variability in bounding box sizes, with each object category exhibiting its own distinct distribution that can be used as prior knowledge in object detection tasks.

\begin{figure*}[t]
    \centering
    \includegraphics[width=0.9\textwidth]{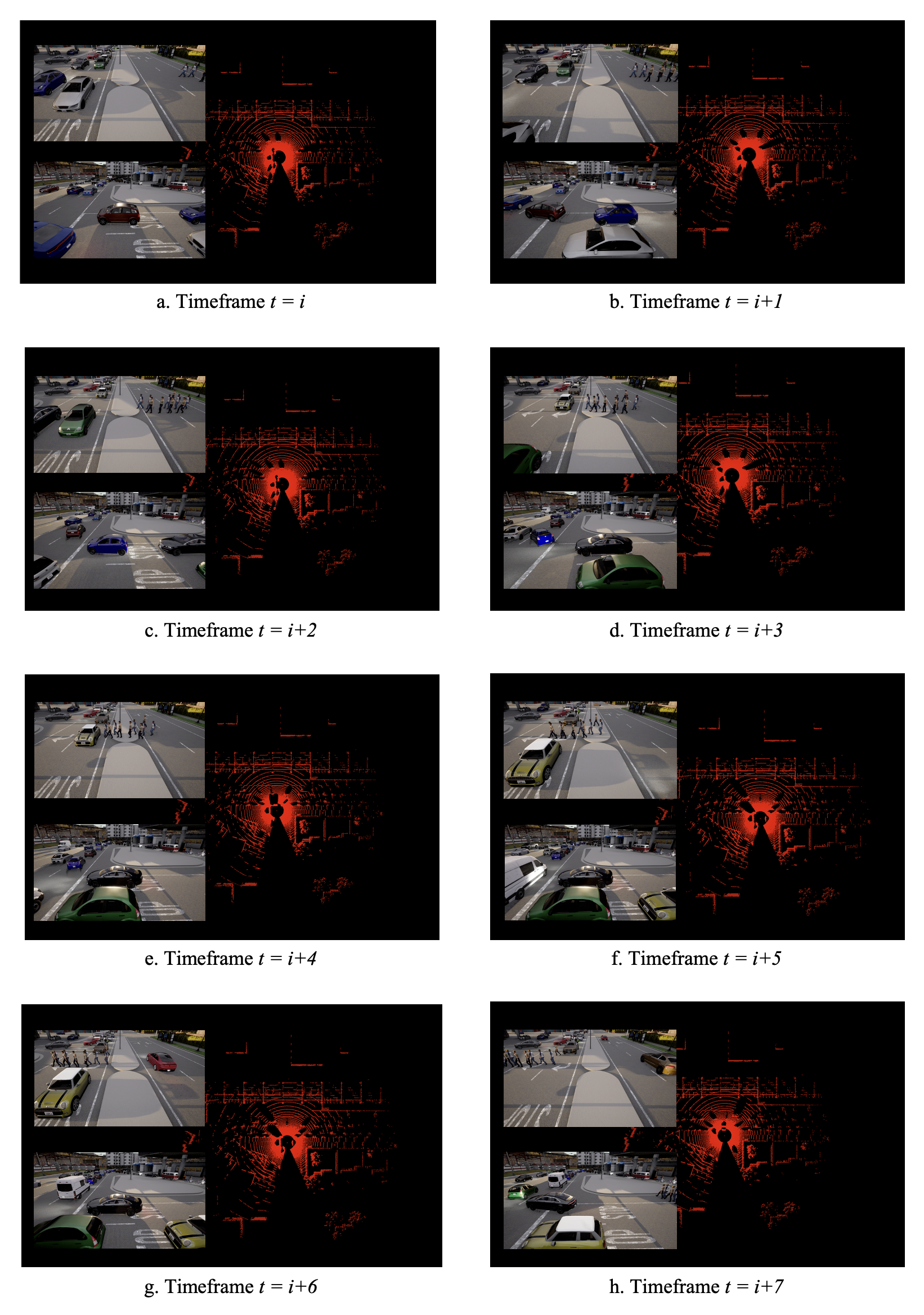}
    \caption{\textbf{Illustration of Multi-Sensor Perception in a Triangular Intersection Across Timeframes}}
    \label{fig:timeframes}
    \vspace{91mm}
\end{figure*}

\subsection{Data Comparision}

Our dataset was compared with other public cooperative perception datasets, and it significantly outperforms its counterparts in terms of the number of intersections, the number of infrastructures, and the overall data volume. Moreover, our dataset is the only one available that supports research on heterogeneous sensor placement.

\subsection{Dataset Visualization}
We visualize a segment of the dataset over a period of time, as shown in Fig. \ref{fig:timeframes}, it includes part of our LiDAR and camera data.

{
    \small
    \bibliographystyle{ieeenat_fullname}
    \bibliography{reference}

\begin{thebibliography}{51}
\providecommand{\natexlab}[1]{#1}
\providecommand{\url}[1]{\texttt{#1}}
\expandafter\ifx\csname urlstyle\endcsname\relax
  \providecommand{\doi}[1]{doi: #1}\else
  \providecommand{\doi}{doi: \begingroup \urlstyle{rm}\Url}\fi

\bibitem[ITE(2015)]{ITE_RSU_Std_v1}
Roadside unit (rsu) standard v1.0.
\newblock Technical report, Institute of Transportation Engineers, 2015.

\bibitem[ITE(2021)]{ITE_RSU_SDR}
Standard development report for roadside unit (rsu).
\newblock Technical report, Institute of Transportation Engineers, 2021.

\bibitem[Barrachina et~al.(2013)Barrachina, Garrido, Fogue, Martinez, Cano, Calafate, and Manzoni]{barrachina2013road}
Javier Barrachina, Piedad Garrido, Manuel Fogue, Francisco~J Martinez, Juan-Carlos Cano, Carlos~T Calafate, and Pietro Manzoni.
\newblock Road side unit deployment: A density-based approach.
\newblock \emph{IEEE Intelligent Transportation Systems Magazine}, 5\penalty0 (3):\penalty0 30--39, 2013.

\bibitem[Bhover et~al.(2017)Bhover, Tugashetti, and Rashinkar]{bhover2017v2x}
Sushma~U Bhover, Anusha Tugashetti, and Pratiksha Rashinkar.
\newblock V2x communication protocol in vanet for co-operative intelligent transportation system.
\newblock In \emph{2017 international conference on innovative mechanisms for industry applications (ICIMIA)}, pages 602--607. IEEE, 2017.

\bibitem[Bresenham(1998)]{bresenham1998algorithm}
Jack~E Bresenham.
\newblock Algorithm for computer control of a digital plotter.
\newblock In \emph{Seminal graphics: pioneering efforts that shaped the field}, pages 1--6. 1998.

\bibitem[Caesar et~al.(2020)Caesar, Bankiti, Lang, Vora, Liong, Xu, Krishnan, Pan, Baldan, and Beijbom]{caesar2020nuscenes}
Holger Caesar, Varun Bankiti, Alex~H Lang, Sourabh Vora, Venice~Erin Liong, Qiang Xu, Anush Krishnan, Yu Pan, Giancarlo Baldan, and Oscar Beijbom.
\newblock nuscenes: A multimodal dataset for autonomous driving.
\newblock In \emph{Proceedings of the IEEE/CVF conference on computer vision and pattern recognition}, pages 11621--11631, 2020.

\bibitem[Cai et~al.()Cai, Jiang, Xu, Zhao, Ma, Liu, and Li]{cai2211analyzing}
X Cai, W Jiang, R Xu, W Zhao, J Ma, S Liu, and Y Li.
\newblock Analyzing infrastructure lidar placement with realistic lidar simulation library. arxiv 2022.
\newblock \emph{arXiv preprint arXiv:2211.15975}.

\bibitem[Cai et~al.(2023)Cai, Zhou, Wang, and Zhao]{cai2023rls}
Xinyu Cai, Yifan Zhou, Chengxi Wang, and Ding Zhao.
\newblock Analyzing infrastructure lidar placement with realistic lidar simulation library.
\newblock \emph{IEEE Transactions on Intelligent Vehicles}, 2023.

\bibitem[Carter et~al.(2006)Carter, Hunter, Zegeer, Stewart, and Huang]{carter2006pedestrian}
Daniel~L Carter, William~W Hunter, Charles~V Zegeer, J~Richard Stewart, and Herman~F Huang.
\newblock Pedestrian and bicyclist intersection safety indices: final report.
\newblock \emph{Federal Highway Administration: McLean, VA, USA}, 2006.

\bibitem[Dosovitskiy et~al.(2017)Dosovitskiy, Ros, Codevilla, Lopez, and Koltun]{dosovitskiy2017carla}
Alexey Dosovitskiy, German Ros, Felipe Codevilla, Antonio Lopez, and Vladlen Koltun.
\newblock Carla: An open urban driving simulator.
\newblock In \emph{Conference on robot learning}, pages 1--16. PMLR, 2017.

\bibitem[Easa et~al.(2020)Easa, Ma, Liu, Yang, and Arkatkar]{easa2020reliability}
Said~M Easa, Yang Ma, Shixu Liu, Yanqun Yang, and Shriniwas Arkatkar.
\newblock Reliability analysis of intersection sight distance at roundabouts.
\newblock \emph{Infrastructures}, 5\penalty0 (8):\penalty0 67, 2020.

\bibitem[Gattis(2010)]{gattis2010guide}
JL Gattis.
\newblock \emph{Guide for the geometric design of driveways}.
\newblock Transportation Research Board, 2010.

\bibitem[Gorrini et~al.(2016)Gorrini, Vizzari, and Bandini]{gorrini2016towards}
Andrea Gorrini, Giuseppe Vizzari, and Stefania Bandini.
\newblock Towards modelling pedestrian-vehicle interactions: Empirical study on urban unsignalized intersection.
\newblock \emph{arXiv preprint arXiv:1610.07892}, 2016.

\bibitem[Guerna et~al.(2022)Guerna, Bitam, and Calafate]{guerna2022roadside}
Abderrahim Guerna, Salim Bitam, and Carlos~T Calafate.
\newblock Roadside unit deployment in internet of vehicles systems: A survey.
\newblock \emph{Sensors}, 22\penalty0 (9):\penalty0 3190, 2022.

\bibitem[Hao et~al.(2024)Hao, Fan, Dai, Zhang, Li, Wang, Yu, Yang, Yuan, and Nie]{hao2024rcooper}
Ruiyang Hao, Siqi Fan, Yingru Dai, Zhenlin Zhang, Chenxi Li, Yuntian Wang, Haibao Yu, Wenxian Yang, Jirui Yuan, and Zaiqing Nie.
\newblock Rcooper: A real-world large-scale dataset for roadside cooperative perception.
\newblock \emph{Proceedings of the IEEE/CVF Conference on Computer Vision and Pattern Recognition (CVPR)}, 2024.

\bibitem[Hao~Xiang(2023)]{hao2023hmvit}
Jiaqi~Ma Hao~Xiang, Runsheng~Xu.
\newblock Hm-vit: Hetero-modal vehicle-to-vehicle cooperative perception with vision transformer.
\newblock \emph{Proceedings of the IEEE/CVF Conference on Computer Vision and Pattern Recognition (CVPR)}, 2023.

\bibitem[Hartley and Zisserman(2003)]{hartley2003multiple}
Richard Hartley and Andrew Zisserman.
\newblock \emph{Multiple view geometry in computer vision}.
\newblock Cambridge university press, 2003.

\bibitem[He et~al.(2016)He, Zhang, Ren, and Sun]{he2016deep}
Kaiming He, Xiangyu Zhang, Shaoqing Ren, and Jian Sun.
\newblock Deep residual learning for image recognition.
\newblock In \emph{Proceedings of the IEEE conference on computer vision and pattern recognition}, pages 770--778, 2016.

\bibitem[Hu et~al.(2021)Hu, Liu, Chitlangia, Agnihotri, and Zhao]{hu2021multi}
Hanjiang Hu, Zuxin Liu, Sharad Chitlangia, Akhil Agnihotri, and Ding Zhao.
\newblock Investigating the impact of multi-lidar placement on object detection for autonomous driving.
\newblock \emph{IEEE Transactions on Intelligent Vehicles}, 2021.

\bibitem[Hu et~al.(2022{\natexlab{a}})Hu, Liu, Chitlangia, Agnihotri, and Zhao]{hu2022investigating}
Hanjiang Hu, Zuxin Liu, Sharad Chitlangia, Akhil Agnihotri, and Ding Zhao.
\newblock Investigating the impact of multi-lidar placement on object detection for autonomous driving.
\newblock In \emph{Proceedings of the IEEE/CVF conference on computer vision and pattern recognition}, pages 2550--2559, 2022{\natexlab{a}}.

\bibitem[Hu et~al.(2022{\natexlab{b}})Hu, Fang, Lei, Zhong, and Chen]{hu2022where2comm}
Yue Hu, Shaoheng Fang, Zixing Lei, Yiqi Zhong, and Siheng Chen.
\newblock Where2comm: Communication-efficient collaborative perception via spatial confidence maps.
\newblock 2022{\natexlab{b}}.

\bibitem[Jiang et~al.(2023)Jiang, Xiang, Cai, Xu, Ma, Li, Lee, and Liu]{jiang2023optimizing}
Wentao Jiang, Hao Xiang, Xinyu Cai, Runsheng Xu, Jiaqi Ma, Yikang Li, Gim~Hee Lee, and Si Liu.
\newblock Optimizing the placement of roadside lidars for autonomous driving.
\newblock In \emph{Proceedings of the IEEE/CVF International Conference on Computer Vision}, pages 18381--18390, 2023.

\bibitem[Kerlow(2009)]{kerlow2009art}
Isaac~V Kerlow.
\newblock \emph{The art of 3D computer animation and effects}.
\newblock John Wiley \& Sons, 2009.

\bibitem[Kim et~al.(2023)Kim, Jo, Yun, Yun, and Park]{kim2023placement}
Tae-Hyeong Kim, Gi-Hwan Jo, Hyeong-Seok Yun, Kyung-Su Yun, and Tae-Hyoung Park.
\newblock Placement method of multiple lidars for roadside infrastructure in urban environments.
\newblock \emph{Sensors}, 23\penalty0 (21):\penalty0 8808, 2023.

\bibitem[Kim et~al.(2024)Kim, Choi, Choi, Ahn, and Hwang]{kim2024enhancing}
Yeonjoo Kim, Byungjoo Choi, Minji Choi, Seunghui Ahn, and Sungjoo Hwang.
\newblock Enhancing pedestrian perceived safety through walking environment modification considering traffic and walking infrastructure.
\newblock \emph{Frontiers in public health}, 11:\penalty0 1326468, 2024.

\bibitem[Li et~al.(2024{\natexlab{a}})Li, Zhou, Wang, and Chen]{li2024multimodal}
X. Li, H. Zhou, Y. Wang, and J. Chen.
\newblock A multi-modal approach for large-scale traffic density estimation.
\newblock \emph{arXiv preprint}, 2401.01454, 2024{\natexlab{a}}.

\bibitem[Li et~al.(2022{\natexlab{a}})Li, Ma, An, Wang, Zhong, Chen, and Feng]{li2022v2xsim}
Yiming Li, Dekun Ma, Ziyan An, Zixun Wang, Yiqi Zhong, Siheng Chen, and Chen Feng.
\newblock V2x-sim: Multi-agent collaborative perception dataset and benchmark for autonomous driving.
\newblock \emph{ArXiv}, Proceedings of the IEEE/CVF Conference on Computer Vision and Pattern Recognition (CVPR), 2022{\natexlab{a}}.

\bibitem[Li et~al.(2022{\natexlab{b}})Li, Ren, Wu, Chen, Feng, and Zhang]{li2022disconet}
Yiming Li, Shunli Ren, Pengxiang Wu, Siheng Chen, Chen Feng, and Wenjun Zhang.
\newblock Learning distilled collaboration graph for multi-agent perception.
\newblock 2022{\natexlab{b}}.

\bibitem[Li et~al.(2024{\natexlab{b}})Li, Hu, Liu, Xu, Huang, and Zhao]{li2024influence}
Ye Li, Hanjiang Hu, Zuxin Liu, Xiaohao Xu, Xiaonan Huang, and Ding Zhao.
\newblock Influence of camera-lidar configuration on 3d object detection for autonomous driving.
\newblock In \emph{2024 IEEE International Conference on Robotics and Automation (ICRA)}, pages 9018--9025. IEEE, 2024{\natexlab{b}}.

\bibitem[Li et~al.(2022{\natexlab{c}})Li, Wang, Li, Xie, Sima, Lu, Yu, and Dai]{li2022bevformer}
Zhiqi Li, Wenhai Wang, Hongyang Li, Enze Xie, Chonghao Sima, Tong Lu, Qiao Yu, and Jifeng Dai.
\newblock Bevformer: Learning bird’s-eye-view representation from multi-camera images via spatiotemporal transformers.
\newblock \emph{Proceedings of the IEEE/CVF Conference on Computer Vision and Pattern Recognition (CVPR)}, 2022{\natexlab{c}}.

\bibitem[Liu et~al.(2022)Liu, Tang, Amini, Yang, Mao, and Rus]{liu2022bevfusion}
Zhijian Liu, Haotian Tang, Alexander Amini, Xinyu Yang, Huizi Mao, and Daniela Rus.
\newblock Bevfusion: Multi-task multi-sensor fusion with unified bird’s-eye view representation.
\newblock \emph{Proceedings of the IEEE/CVF Conference on Computer Vision and Pattern Recognition (CVPR)}, 2022.

\bibitem[Lu et~al.(2023)Lu, Li, Liu, Dianati, Feng, Chen, and Wang]{lu2023coalign}
Yifan Lu, Quanhao Li, Baoan Liu, Mehrdad Dianati, Chen Feng, Siheng Chen, and Yanfeng Wang.
\newblock Robust collaborative 3d object detection in presence of pose errors.
\newblock 2023.

\bibitem[Lu et~al.(2024)Lu, Hu, Zhong, Wang, Wang, and Chen]{lu2024eheal}
Yifan Lu, Yue Hu, Yiqi Zhong, Dequan Wang, Yanfeng Wang, and Siheng Chen.
\newblock An extensible framework for open heterogeneous collaborative perception.
\newblock 2024.

\bibitem[Luo et~al.(2023)Luo, Sun, Zhang, Yu, and Liu]{luo2023seip}
Jie Luo, Zhi Sun, Hao Zhang, Zhenyu Yu, and Feng Liu.
\newblock Seip: Simulation-based design and evaluation of infrastructure-based collective perception.
\newblock \emph{arXiv preprint arXiv:2305.17892}, 2023.

\bibitem[Ma et~al.(2021)Ma, Liu, and Li]{ma2021perception}
Tao Ma, Zhizheng Liu, and Yikang Li.
\newblock Perception entropy: A metric for multiple sensors configuration evaluation and design.
\newblock \emph{arXiv preprint arXiv:2104.06615}, 2021.

\bibitem[McGhee(1998)]{mcghee1998traffic}
C.~C. McGhee.
\newblock Traffic flow theory and characteristics: A review and evaluation of available models.
\newblock Technical report, Virginia Transportation Research Council, 1998.

\bibitem[McMahon(2002)]{mcmahon2002analysis}
Patrick~J McMahon.
\newblock \emph{An analysis of factors contributing to" walking along roadway" crashes research study and guidelines for sidewalks and walkways}.
\newblock DIANE Publishing, 2002.

\bibitem[Philion and Fidler(2020)]{philion2020lift}
Jonah Philion and Sanja Fidler.
\newblock Lift, splat, shoot: Encoding images from arbitrary camera rigs by implicitly unprojecting to 3d.
\newblock In \emph{Computer Vision--ECCV 2020: 16th European Conference, Glasgow, UK, August 23--28, 2020, Proceedings, Part XIV 16}, pages 194--210. Springer, 2020.

\bibitem[Qu et~al.(2023)Qu, Huang, and Suo]{qu2023seip}
Ao Qu, Xuhuan Huang, and Dajiang Suo.
\newblock Seip: Simulation-based design and evaluation of infrastructure-based collective perception.
\newblock In \emph{2023 IEEE 26th International Conference on Intelligent Transportation Systems (ITSC)}, pages 3871--3878. IEEE, 2023.

\bibitem[Sultana et~al.(2014)Sultana, Qin, Chitturi, and Noyce]{sultana2014analysis}
Most~Afia Sultana, Xiao Qin, Madhav Chitturi, and David~A Noyce.
\newblock Analysis of safety effects of traffic, geometric, and access parameters on truck arterial corridors.
\newblock \emph{Transportation research record}, 2404\penalty0 (1):\penalty0 68--76, 2014.

\bibitem[Tan and Le(2019)]{tan2019efficientnet}
Mingxing Tan and Quoc Le.
\newblock Efficientnet: Rethinking model scaling for convolutional neural networks.
\newblock In \emph{International conference on machine learning}, pages 6105--6114. PMLR, 2019.

\bibitem[TH et~al.(2023)TH, GH, HS, KS, and TH]{kim2023lidar}
Kim TH, Jo GH, Yun HS, Yun KS, and Park TH.
\newblock Placement method of multiple lidars for roadside infrastructure in urban environments.
\newblock \emph{Sensors}, 2023.

\bibitem[Vijay et~al.(2021)Vijay, Cherian, Riah, de~Boer, and Choudhury]{vijay2021optimalplacement}
Roshan Vijay, Jim Cherian, Rachid Riah, Niels de Boer, and Apratim Choudhury.
\newblock Optimal placement of roadside infrastructure sensors towards safer autonomous vehicle deployments, 2021.

\bibitem[Wang et~al.(2020)Wang, Manivasagam, Liang, Yang, Zeng, Tu, and Urtasun]{wang2020v2vnet}
Tsun-Hsuan Wang, Sivabalan Manivasagam, Ming Liang, Bin Yang, Wenyuan Zeng, James Tu, and Raquel Urtasun.
\newblock V2vnet: Vehicle-to-vehicle communication for joint perception and prediction.
\newblock 2020.

\bibitem[Xiang et~al.(2024)Xiang, Zheng, Xia, Xu, Gao, Zhou, Han, Ji, Li, Meng, Jin, Lei, Ma, He, Ma, Yuan, Zhao, and Ma]{xiang2024v2xreal}
Hao Xiang, Zhaoliang Zheng, Xin Xia, Runsheng Xu, Letian Gao, Zewei Zhou, Xu Han, Xinkai Ji, Mingxi Li, Zonglin Meng, Li Jin, Mingyue Lei, Zhaoyang Ma, Zihang He, Haoxuan Ma, Yunshuang Yuan, Yingqian Zhao, and Jiaqi Ma.
\newblock V2x-real: a large-scale dataset for vehicle-to-everything cooperative perception.
\newblock 2024.

\bibitem[Xu et~al.(2021)Xu, Xiang, Xia, Han, Li, and Ma]{xu2021opv2v}
Runsheng Xu, Hao Xiang, Xin Xia, Xu Han, Jinlong Li, and Jiaqi Ma.
\newblock Opv2v: An open benchmark dataset and fusion pipeline for perception with vehicle-to-vehicle communication.
\newblock 2021.

\bibitem[Xu et~al.(2022)Xu, Xiang, Tu, Xia, Yang, and Ma]{xu2022v2xvit}
Runsheng Xu, Hao Xiang, Zhengzhong Tu, Xin Xia, Ming-Hsuan Yang, and Jiaqi Ma.
\newblock V2x-vit: Vehicle-to-everything cooperative perception with vision transformer.
\newblock 2022.

\bibitem[Xu et~al.(2023)Xu, Xia, Li, Li, Zhang, Tu, Meng, Xiang, Dong, Song, Yu, Zhou, and Ma]{xu2023v2v4real}
Runsheng Xu, Xin Xia, Jinlong Li, Hanzhao Li, Shuo Zhang, Zhengzhong Tu, Zonglin Meng, Hao Xiang, Xiaoyu Dong, Rui Song, Hongkai Yu, Bolei Zhou, and Jiaqi Ma.
\newblock V2v4real: A real-world large-scale dataset for vehicle-to-vehicle cooperative perception.
\newblock \emph{Proceedings of the IEEE/CVF Conference on Computer Vision and Pattern Recognition (CVPR)}, 2023.

\bibitem[Yu et~al.(2022)Yu, Luo, Shu, Huo, Yang, Shi, Guo, Li, Hu, Yuan, and Nie]{yu2022dairv2x}
Haibao Yu, Yizhen Luo, Mao Shu, Yiyi Huo, Zebang Yang, Yifeng Shi, Zhenglong Guo, Hanyu Li, Xing Hu, Jirui Yuan, and Zaiqing Nie.
\newblock Dair-v2x: A benchmark dataset for vehicle-infrastructure cooperative perception.
\newblock \emph{Proceedings of the IEEE/CVF Conference on Computer Vision and Pattern Recognition (CVPR)}, 2022.

\bibitem[Yuan et~al.(2022)Yuan, Cheng, and Sester]{yuan2022keypoints}
Yunshuang Yuan, Hao Cheng, and Monika Sester.
\newblock Keypoints-based deep feature fusion for cooperative vehicle detection of autonomous driving.
\newblock \emph{IEEE Robotics and Automation Letters}, 7\penalty0 (2):\penalty0 3054--3061, 2022.

\bibitem[Zhou et~al.(2024)Zhou, Xiang, Zheng, Zhao, Lei, Zhang, Cai, Liu, Liu, Bajji, Pham, Xia, Huang, Zhou, and Ma]{zhou2024v2xpnp}
Zewei Zhou, Hao Xiang, Zhaoliang Zheng, Seth~Z. Zhao, Mingyue Lei, Yun Zhang, Tianhui Cai, Xinyi Liu, Johnson Liu, Maheswari Bajji, Jacob Pham, Xin Xia, Zhiyu Huang, Bolei Zhou, and Jiaqi Ma.
\newblock V2xpnp: Vehicle-to-everything spatio-temporal fusion for multi-agent perception and prediction.
\newblock 2024.

\end{thebibliography}
}

\end{document}